# DETEKSI SAMPAH DI PERMUKAAN DAN DALAM PERAIRAN PADA OBJEK VIDEO DENGAN METODE *ROBUST AND EFFICIENT POST-PROCESSING* DAN *TUBELET-LEVEL BOUNDING BOX LINKING*

| Bryan Tjandra | Made Swastika Nata Negara | Nyoo Steven Christopher H. |
|---|---|---|
| *Faculty of Computer Science* | *Faculty of Computer Science* | *Faculty of Computer Science* |
| *University of Indonesia* | *University of Indonesia* | *University of Indonesia* |
| Depok, Indonesia | Depok, Indonesia | Depok, Indonesia |
| bryantjandra@ristek.cs.ui.ac.id | swastikanata@ristek.cs.ui.ac.id | nyoo@ristek.cs.ui.ac |

## ABSTRAK

Indonesia sebagai negara maritim memiliki wilayah yang didominasi oleh perairan. Pengelolaan sampah yang belum efektif berakhir pada banyaknya sampah di perairan Indonesia yang menyebabkan berbagai macam masalah. Pengembangan robot pengumpul sampah otomatis dapat menjadi solusi untuk mengatasi masalah tersebut. Robot tersebut membutuhkan sistem yang dapat melakukan deteksi pada objek bergerak seperti pada video. Namun, penggunaan metode deteksi objek *naive* pada video memiliki limitasi, terutama ketika fokus gambar berkurang dan objek target tertutup oleh objek lain. Kontribusi makalah ini memberikan penjelasan mengenai metode yang dapat diterapkan untuk melakukan deteksi objek video pada robot pengumpul sampah otomatis. Penelitian ini menggunakan *model* YOLOv5 dan metode *robust & efficient post processing* (REPP) serta *tubelet-level bounding box linking* pada dataset FloW dan Roboflow. Penggabungan metode ini meningkatkan performa deteksi objek secara *naive* dari YOLOv5 dengan mempertimbangkan hasil deteksi pada *frame-frame* yang bersebelahan. Diperoleh hasil bahwa tahap *post processing* dan *tubelet-level bounding box linking* dapat meningkatkan kualitas hasil deteksi, yaitu sekitar 3% lebih baik dibandingkan YOLOv5 saja. Penggunaan metode ini berpotensi untuk mendeteksi sampah di permukaan perairan dan diaplikasikan pada suatu robot pengumpul sampah berbasis citra secara *real time*. Implementasi sistem ini diharapkan dapat memperbaiki kerusakan yang telah disebabkan oleh sampah pada masa lampau dan membuat Indonesia memiliki sistem pengelolaan sampah yang lebih baik pada masa mendatang.

# BAB 1
# PENDAHULUAN

## 1.1. Latar Belakang

Indonesia dikenal sebagai negara maritim terbesar di dunia. Menurut Indonesia Baik (n.d.), sebanyak 62% wilayah Indonesia ditutupi oleh laut dan perairan. Ironisnya, menurut data Kementerian Lingkungan Hidup dan Kehutanan (KLHK), pada tahun 2020, berat sampah di perairan Indonesia diperkirakan mencapai 5.75 juta ton. Menurut UNEP (2021), sampah ini dapat menimbulkan berbagai macam masalah seperti masalah air bersih, sanitasi, dan terancamnya biota air.

Berdasarkan data BPS 2021, DKI Jakarta merupakan salah satu kota dengan penyumbang sampah terbanyak dengan total 7.2 ribu ton per hari. Ditambah dengan adanya pandemi COVID19, Indonesia saat ini sedang memasuki fase kritis dalam pemulihan kesehatan masyarakat yang mengguncang negara ini. Namun, banyaknya sampah di perairan Indonesia, khususnya di sungai dan lautan sama sekali tidak mendukung pemulihan kesehatan tersebut.

Pemerintah Indonesia sudah mengeluarkan beberapa kebijakan dan solusi untuk menangani masalah ini, seperti membangun sistem pengelolaan air yang terpadu dan menegakkan peraturan terkait pencemaran air. Namun, volume sampah yang dihasilkan setiap harinya terlalu besar, sehingga dibutuhkan solusi lebih lanjut. Menurut Lubis (2019), pengumpulan sampah di Indonesia masih dilakukan secara manual. Hal tersebut tidak efisien dan *scalable* untuk perairan Indonesia yang sangat luas.

Besarnya perbandingan jumlah sampah dan luas perairan Indonesia membuat pengembangan mesin pengumpul sampah otomatis seperti robot tanpa awak yang dapat mendeteksi sampah di permukaan ataupun dalam perairan dapat menjadi solusi. Orca-Tech, sebuah perusahaan di China, juga pernah menerapkan solusi ini yang mampu menampung sampah sebanyak 25-50 kg dengan hasil efisiensi 7 kali lebih cepat daripada manusia. Mesin tersebut perlu dibekali dengan sistem yang dapat melakukan deteksi pada objek bergerak pada video, khususnya pada sampah plastik karena sulit terurai. Dengan adanya sistem yang diimplementasikan di robot tanpa awak ini dapat menghemat biaya yang dikeluarkan dan kebutuhan *resource* yang lebih minim. Namun, deteksi pada video lebih sulit dilakukan dibandingkan pada gambar karena rawan terjadi hilangnya fokus gambar dan tertutupnya objek target.

Untuk menangani masalah tersebut, diusulkan sebuah kerangka model yang dapat melakukan deteksi sampah di permukaan air pada objek video dengan menggabungkan metode objek deteksi, *robust & efficient post-processing*, serta *tubelet-level bounding box linking*. Keberhasilan dikembangkannya kerangka ini diharapkan dapat meningkatkan kualitas pengelolaan sampah di permukaan air yang sejalan dengan tujuan Indonesia untuk pulih lebih cepat dan bangkit lebih kuat.

## 1.2. Tujuan dan Manfaat

Penelitian ini bertujuan untuk memaparkan secara detail metode yang dapat digunakan untuk mendeteksi sampah di permukaan perairan pada objek video. Selain itu, penelitian ini memiliki beberapa manfaat seperti:
1. Mengimplementasikan sistem deteksi sampah di permukaan perairan Indonesia secara otomatis yang berasal dari data video.

2. Mendukung upaya persediaan akses air bersih dan kelestarian biota air, sejalan dengan *Sustainable Development Goals* (SDG) ke-6 dan ke-14.
3. Meminimalisasi biaya dan kebutuhan sumber daya manusia yang diperlukan untuk pembersihan sampah di perairan.

**1.3. Batasan Penelitian**

Metode yang dipaparkan dalam makalah memiliki beberapa batasan sebagai berikut:
1. Seluruh dataset gambar dan video memiliki latar belakang perairan dan bukan latar belakang lainnya.
2. Sampah yang dideteksi hanya berupa plastik.
3. Sampah yang terdapat pada dataset berkisar antara 0-30 sampah per frame

# BAB 2
# STUDI LITERATUR

## 2.1. Deteksi Objek

Deteksi objek merupakan salah satu cabang *computer vision* yang bertujuan untuk mengenali, memprediksi objek, dan membuat *bounding box*. Dengan memanfaatkan metode kecerdasan buatan, informasi mengenai objek tersebut dapat diekstrak untuk berbagai manfaat. Deteksi objek dilakukan dengan menemukan lokasi objek tertentu dalam suatu gambar, kemudian menggambarkan kotak pembatas atau *bounding box* yang mengelilingi objek tersebut (Nufus, N et.al, 2021). Salah satu algoritma untuk deteksi objek adalah YOLOv5, yang terkenal akan kecepatan dan akurasinya.

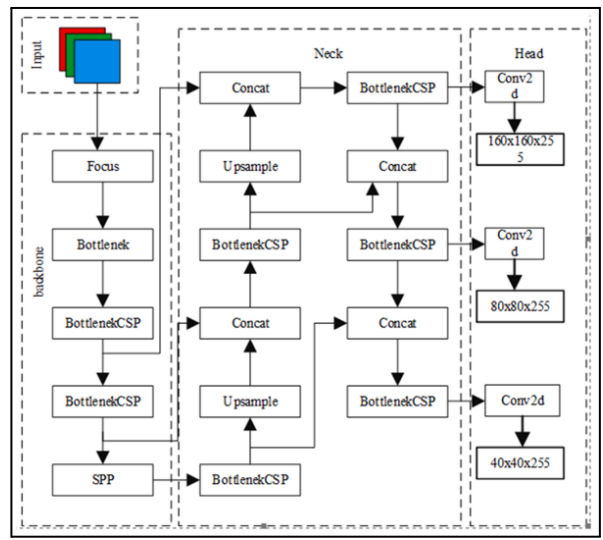

**Gambar 2.1.** Arsitektur YOLOv5 (Sumber:https://doi.org/10.1016/j.aquaeng.2022.102273)

YOLOv5 menggabungkan *cross stage partial network* (CSPNet) dengan Darknet sebagai *backbone*. Metode ini mengurangi ukuran model, sehingga dapat memastikan kecepatan inferensi dan akurasi yang lebih baik. YOLOv5 juga menerapkan jaringan PANet sebagai *neck* untuk meningkatkan arus informasi. Lalu, lapisan YOLO sebagai *head* menghasilkan 3 peta fitur dengan ukuran berbeda (18 × 18, 36 × 36, 72 × 72) untuk mencapai prediksi multi-skala. Hal ini memungkinkan model menangani objek kecil, sedang, dan besar (Renjie Xu et.al, 2021).

## 2.2. Deteksi Objek Video

Inti dari deteksi objek video hampir serupa dengan deteksi objek pada gambar, perbedaannya adalah objek yang dideteksi terdapat dalam *frame-frame* dari video. Selain itu, apabila deteksi objek pada gambar menghasilkan *bounding box*, deteksi objek pada video menghasilkan *tubelet*, yaitu *bounding box* terurut yang didapat dari masing-masing *frame*. Deteksi objek video cenderung lebih sulit untuk dilakukan karena tingginya variansi kemunculan objek pada setiap *frame* dan penurunan kualitas gambar yang dapat terjadi pada setiap *frame*. Namun, terdapat sisi positif di mana deteksi objek video dapat memanfaatkan korelasi spasial dan temporal antar-*sequence* untuk mendapatkan hasil deteksi yang baik. Dengan demikian, cara penggabungan fitur antar-*sequence* menjadi hal yang penting dalam deteksi objek video (Kang et al., 2017).

## 2.3. *Robust and Efficient Post-Processing*

*Robust and Efficient Post-Processing* (REPP) merupakan metode *learning-based similarity* yang mengevaluasi *bounding box* objek antar urutan gambar yang dapat dimanfaatkan dalam deteksi objek video. Metode ini menggunakan dasar dari hasil *still-image detector* seperti YOLOv5. Untuk semua pasangan deteksi dari *frame* berurutan (*t* sampai *t* + *i*), dibangun satu set fitur berdasarkan lokasi, geometri, tampilan, dan semantiknya untuk memprediksi *linking (similarity) score*.

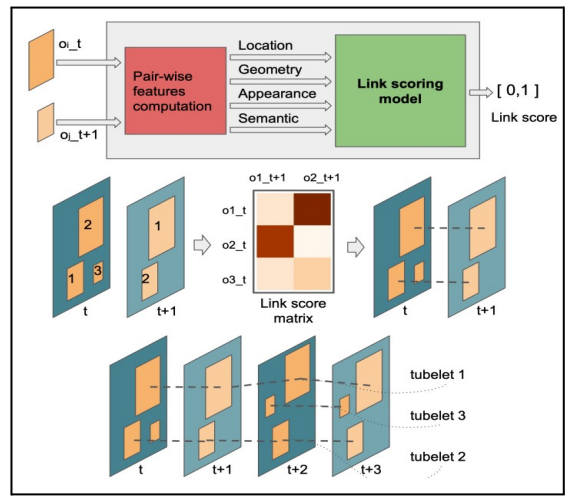

**Gambar 2.2.** Penautan deteksi objek dengan REPP (Sumber:https://arxiv.org/abs/2009.11050)

Skor yang tinggi digunakan untuk menghubungkan deteksi ke urutan berikutnya dengan membangun satu set *tubelet*. Pada setiap gambar dalam *tubelet*, akan dilakukan perbaikan *confidence score* dan koordinat *bounding box* untuk menciptakan *bounding box* yang lebih tidak fluktuatif (Alberto Sabater et al., 2020).

### 2.4. *Tubelet-Level Bounding Box Linking*

*Tubelet-level bounding box linking* merupakan metode yang digunakan untuk memperbaiki deteksi objek yang tidak tepat dan meningkatkan *recall score*. Berbeda dengan REPP yang hanya menghubungkan *bounding box* pada dua *frame* yang bersebelahan, *tubelet-level bounding box linking* menghubungkan akhir dari suatu *tubelet* ke awal dari *tubelet* yang lain.

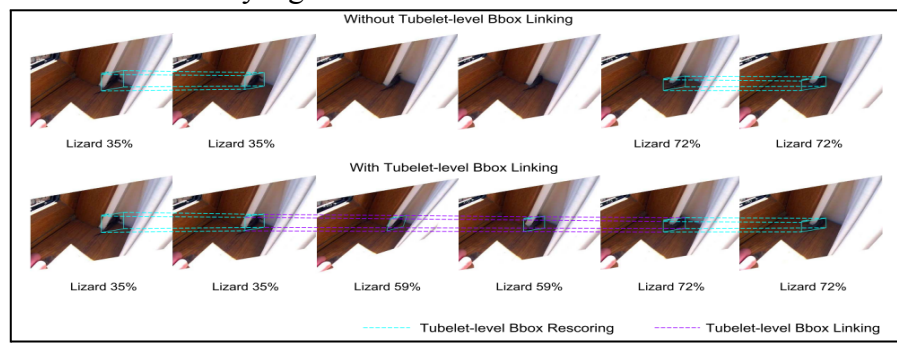

**Gambar 2.3.** Ilustrasi *tubelet-level bbox linking* (Sumber:www.scitepress.org/Papers/2019/72600/72600.pdf)

Suatu nilai batasan dipilih untuk menentukan interval maksimal dari dua *tubelet* yang akan dipasangkan. *Tubelets* dengan interval temporal yang lebih besar dari nilai batas tersebut dianggap tidak berhubungan. Sebaliknya, jika akhir dari suatu *tubelet* dan awal dari *tubelet* yang lain dinilai berhubungan, kedua *tubelet* tersebut dianggap sebagai hasil deteksi objek yang sama. Jika kedua *tubelet* saling terhubung, nilai dari *confidence score* untuk objek di antara kedua tubelet yang tidak terdeteksi adalah rata-rata *confidence score* kedua *tubelet* tersebut (Belhassen, H et al., 2019).

# BAB 3
# METODOLOGI

## 3.1. Dataset

Dataset yang digunakan dalam penelitian ini adalah dataset FloW dan Roboflow, yaitu sebuah dataset yang ditujukan untuk mendeteksi sampah di permukaan dan dalam perairan. Dataset FloW terdiri atas 2000 gambar yang diambil menggunakan kapal tak berawak, sedangkan Roboflow terdiri atas 1000 gambar sampah di dalam air. Dataset ini memiliki lebih dari 8000 sampah di permukaan perairan yang memiliki anotasi. Selain gambar, terdapat pula 43 video pendek yang belum diberikan anotasi.

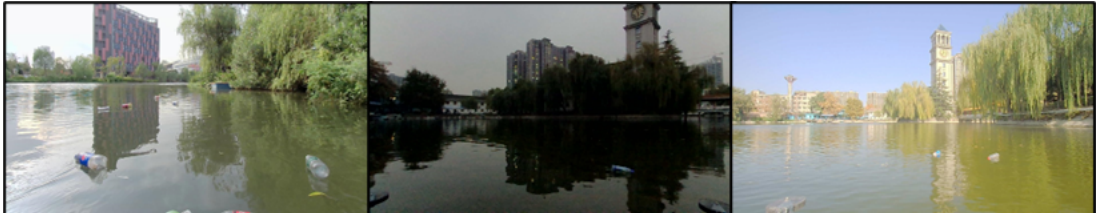

**Gambar 3.1.1.** Sampel gambar sampah di permukaan air

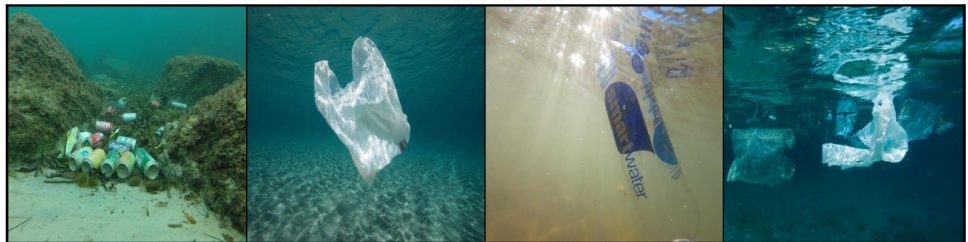

**Gambar 3.1.2.** Sampel gambar sampah di dalam air

| **Jenis** | *Training* | *Validation* | *Testing* |
|---|---|---|---|
| **Data Sampah Permukaan Air** | 2000 Gambar | 10 Video | 25 Video |
| **Data Sampah Bawah Air** | 1000 Gambar | 3 Video | 5 Video |

**Tabel 3.1.** Perbandingan jumlah data pada *training*, *validation*, dan *test set*

*Validation set* dipilih berdasarkan latar belakang waktu (pagi dan siang), kondisi air (kotor dan jernih), serta tingkat resolusi gambar (tinggi dan rendah). *Validation set* dipilih berdasarkan latar belakang waktu (pagi dan siang), kondisi air (kotor dan jernih), serta tingkat resolusi gambar (tinggi dan rendah).

## 3.2. Metode Eksperimen

Secara garis besar, eksperimen yang akan tim peneliti lakukan digambarkan melalui skema diagram di bawah ini.

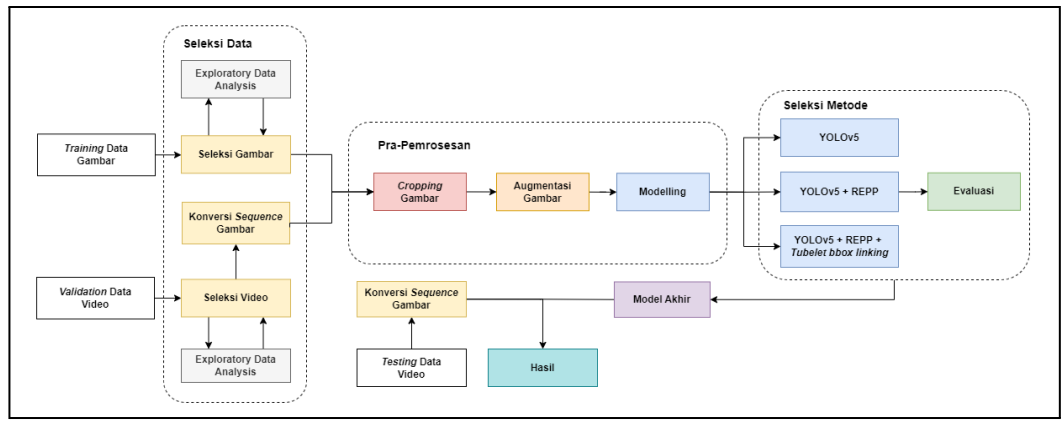

**Gambar 3.2.** Skema diagram alur eskperimen.

Pada awalnya, video dan gambar dianalisis terlebih dahulu untuk menyeleksi data yang relevan. Data video kemudian dikonversi terlebih dahulu ke *image sequence* dengan spesifikasi 20 fps. *Image sequence* tersebut dianotasi secara *semi-supervised* untuk menghasilkan *ground truth*. *Preprocessing* yang dilakukan pada gambar berupa *cropping* dengan resolusi 1280x720 px untuk sampah di permukaan air dan 416x416 untuk sampah di dalam air. Kemudian, dilakukan augmentasi gambar dengan augmentasi bawaan dari YOLOv5 seperti *flip, blur, to gray*, dan CLACHE.

Eksperimen deteksi objek video dilakukan dengan membandingkan hasil penggunaan model deteksi objek secara *naive* dan model deteksi objek yang didukung metode *post-processing* serta *tubelet-level bounding box linking*. Model yang digunakan sebagai *baseline* merupakan hasil *fine-tuning* model YOLOv5 dengan iterasi 20 *epochs*, *learning rate* 0.01, dan *batch size* 64.

Hasil prediksi model YOLOv5 dijadikan sebagai *input* ke REPP dan *tubelet-level bounding box linking*. Metode ini digunakan untuk mempertimbangkan hasil prediksi dari gambar ke-*t* sampai gambar ke-(*t+i*) untuk menghasilkan prediksi baru dan mengurangi prediksi *false positive*. Setelah itu, hasil prediksi tersebut dievaluasi menggunakan *metrics mean average precision* (mAP50 dan mAP50-95) untuk melihat perbandingan *improvement* dari kedua metode tersebut. mAP dipilih untuk memaksimalkan hasil *precision* dan *recall* dimana keduanya sama-sama penting.

# BAB 4
# HASIL DAN PEMBAHASAN

## 4.1 Hasil Pengujian

Hasil dari *fine tuning* model YOLOv5 ditunjukkan pada grafik berikut:

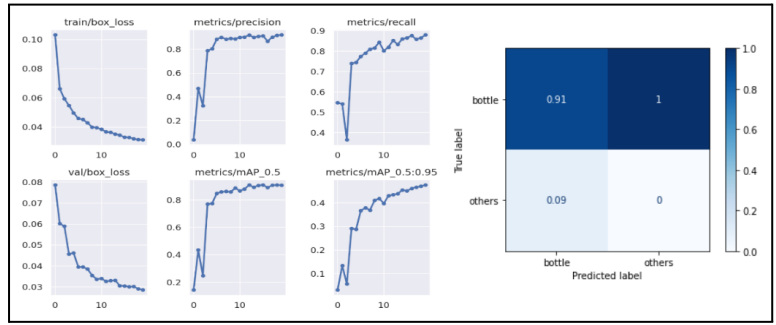

**Gambar 3.3.1.** Hasil *fine-tuning* YOLOv5 pada *image* sampah di permukaan air.

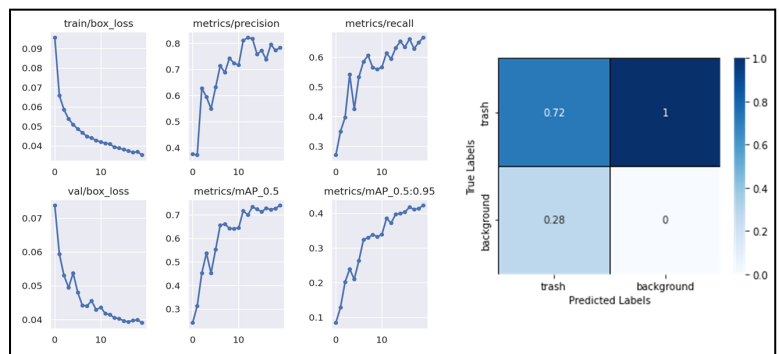

**Gambar 3.3.2.** Hasil *fine-tuning* YOLOv5 pada *image* sampah di bawah air.

Hasil *fine-tuning* model pada *epoch* terakhir memperoleh *loss* pada gambar permukaan dan dalam air berturut turut sebesar 0.03 dan 0.04, serta mAP0.5 pada *validation set* meningkat dari 0.14 hingga 0.88 dan 0.21 hingga 0.79. *False positive* yang terdapat pada model berturut-turut sebesar 0.09 dan 0.28 dari validasi. Hasil dari *fine-tuning* ini menunjukkan *reliability* dari model yang digunakan sebagai *baseline* deteksi objek. Berikut beberapa hasil prediksi dari model *baseline* terhadap gambar validasi.

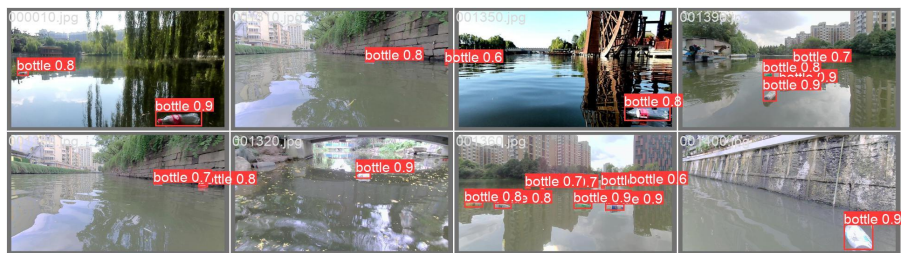

**Gambar 3.4.1.** Hasil prediksi *baseline* terhadap *validation image* sampah di permukaan air.

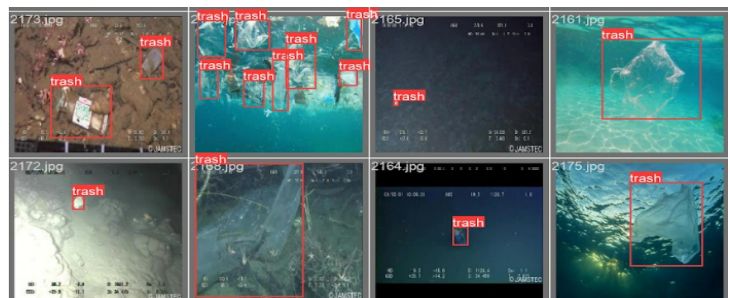

**Gambar 3.4.2.** Hasil prediksi *baseline* terhadap *validation image* sampah di dalam air

Skor mAP dari YOLOv5 sebelum dan setelah dilakukan metode *robust & efficient post-processing* dan *tubelet-level bounding box linking* ditunjukkan pada tabel berikut.

| Model | mAP50 Score | mAP50-95 Score |
|---|---|---|
| YOLOv5 (*baseline*) | 0.884 | 0.4526 |
| YOLOv5 + REPP | 0.904 | 0.4724 |
| YOLOv5 + REPP + *Bbox linking* | 0.915 | 0.4817 |

**Tabel 3.2.1.** *mAP score* dari masing-masing *approach* untuk gambar sampah di permukaan air.

| Model | mAP50 Score | mAP50-95 Score |
|---|---|---|
| YOLOv5 (*baseline*) | 0.795 | 0.4232 |
| YOLOv5 + REPP | 0.813 | 0.4314 |
| YOLOv5 + REPP + *Bbox linking* | 0.828 | 0.4476 |

**Tabel 3.2.2.** *mAP score* dari masing-masing *approach* untuk gambar sampah di dalam air.

Hasil evaluasi model YOLOv5 dengan metode REPP dan *tubelets bbox linking* memperoleh peningkatan mAP sekitar 3% dibandingkan model objek deteksi *naive*.

### 4.2 Analisis

Dengan metode deteksi objek secara *naive*, model dapat memperoleh skor mAP yang tinggi. Namun, masih terdapat beberapa *false positive* dan objek yang tidak terdeteksi. Hal tersebut disebabkan oleh sifat metode ini yang tidak mempertimbangkan *sequence* sebelumnya untuk prediksi *bounding box* pada *sequence* saat ini.

Setelah menerapkan *robust & efficient post-processing* dan *tubelet-level bounding box linking*, *false positive* dapat diperbaiki dan *missing object* dapat dideteksi. Penggunaan REPP bertujuan untuk *rescoring confidence score* dan menggunakan *linking score* untuk menghilangkan kesalahan deteksi (*false positive*), tetapi REPP tidak menerapkan *bounding box linking*. Oleh karena itu, *tubelet-level bounding box linking* digunakan untuk menghasilkan interpolasi linear dari akhir suatu *tubelet* ke awal *tubelet* yang lain untuk mendeteksi *missing object*. Berikut beberapa demonstrasi bagaimana metode yang diusulkan dapat meningkatkan performa deteksi.

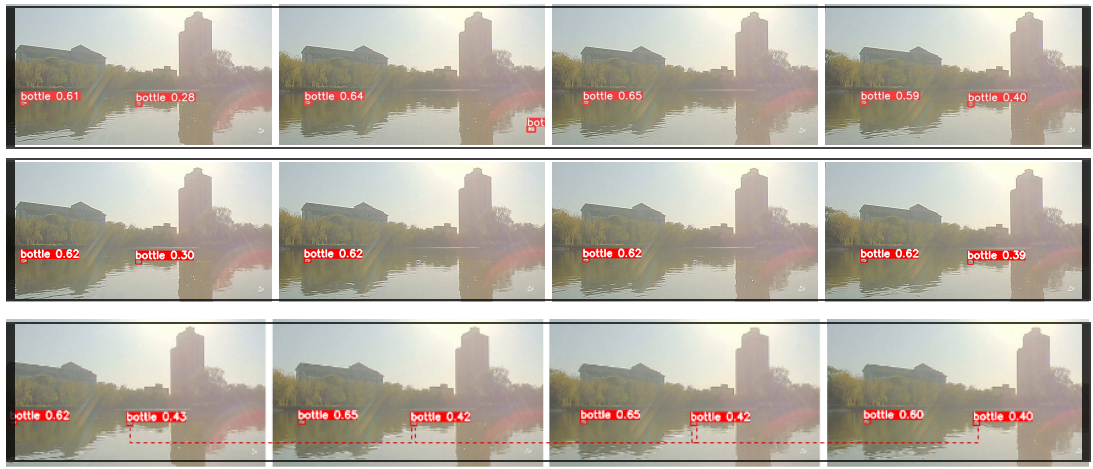

**Gambar 3.5.1.** Peningkatan performa setelah mengaplikasikan REPP dan *bounding box linking* pada sampah permukaan air. (a) Hasil deteksi YOLOv5 *naive*. (b) Hasil deteksi YOLOv5 dengan REPP. (c) Hasil deteksi YOLOv5 dengan REPP dan *bounding box linking*.

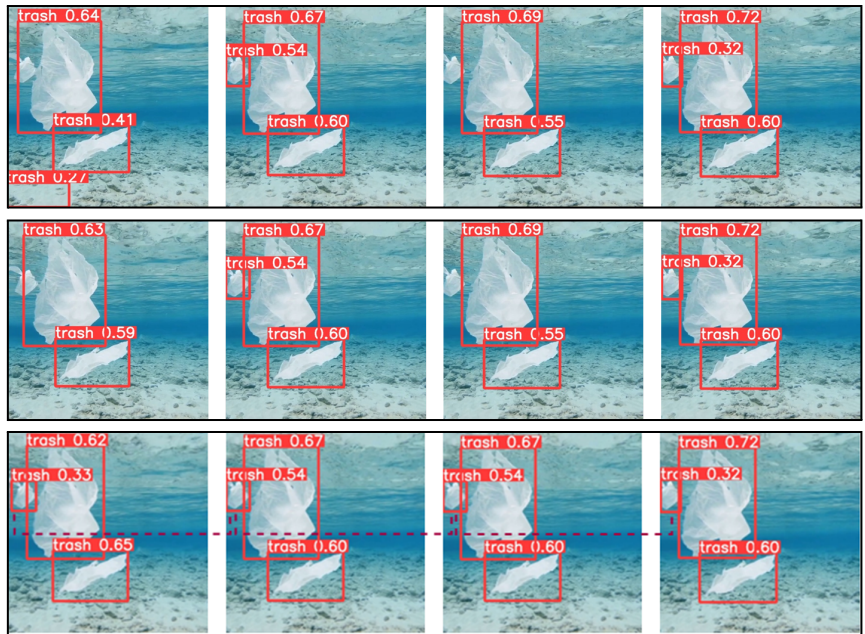

**Gambar 3.5.2.** Peningkatan performa setelah mengaplikasikan REPP dan *bounding box linking* pada sampah di dalam air. (a) Hasil deteksi YOLOv5 *naive*. (b) Hasil deteksi YOLOv5 dengan REPP. (c) Hasil deteksi YOLOv5 dengan REPP dan *bounding box linking*.

Deteksi dari model YOLOv5 *naive* memiliki beberapa hasil *false positive* dan objek tidak terdeteksi pada frame kedua dan ketiga pada gambar (a). Metode REPP dan model YOLOv5 mampu mengurangi *false positive* dan meningkatkan *confidence score* pada anotasi gambar (b). Penambahan metode *tubelet bounding box linking* dapat mengurangi objek tidak terdeteksi dengan mempertimbangkan *frame* sebelum dan sesudahnya, seperti terlihat pada gambar (c).

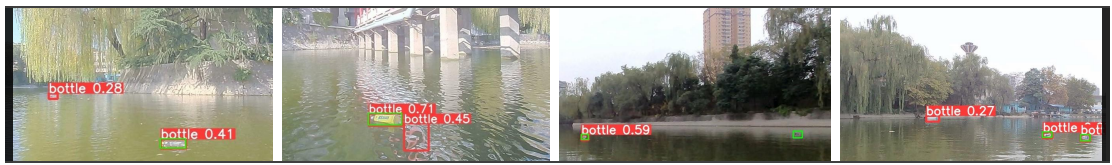

**Gambar 3.6.** Kesalahan deteksi meskipun sudah mengaplikasikan REPP dan bounding box linking (Keterangan: box hijau menunjukkan ground truth, sedangkan box merah menunjukkan hasil prediksi)

Metode YOLOv5, REPP, dan *bounding box linking* dapat meningkatkan performa deteksi objek video dibandingkan dengan deteksi objek secara *naive*. Namun, dikarenakan faktor resolusi rendah, objek yang memiliki kemiripan dengan sampah, dan intensitas cahaya tinggi, terdapat beberapa prediksi yang salah, seperti yang dilihat pada gambar 3.6.

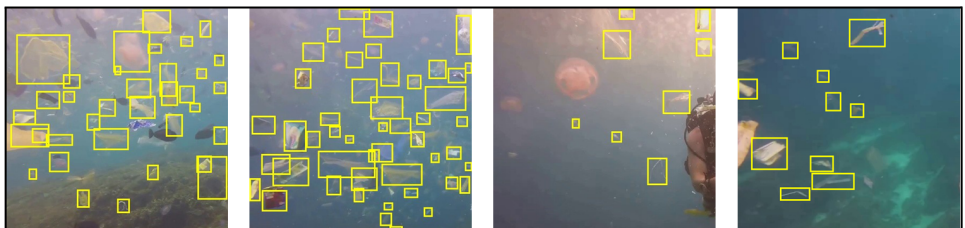

**Gambar 3.6.** Kesalahan deteksi meskipun sudah mengaplikasikan REPP dan bounding box linking (Keterangan: box hijau menunjukkan ground truth, sedangkan box merah menunjukkan hasil prediksi)

Metode ini mampu membedakan antara objek sampah dan air, serta objek lainnya seperti makhluk laut dengan hasil deteksi yang akurat. Selain itu, hasil deteksi juga menunjukkan bahwa metode ini dapat digunakan untuk mendeteksi jenis sampah plastik dengan jangkauan mencapai 30 sampah pada 1 frame. Dengan ini, model YOLOv5 dengan REPP dan *bounding box linking* dapat menjadi solusi yang efektif dalam membantu mengurangi jumlah sampah di dalam air dan menjaga lingkungan laut yang bersih dan sehat.

# BAB 4
# PENUTUP

## 4.1 Kesimpulan

Implementasi *deep learning* dengan memanfaatkan *pretrained model* YOLOv5 dapat melakukan deteksi sampah di permukaan perairan dengan hasil yang cukup baik. Dari eksperimen yang telah dilakukan, disimpulkan bahwa tahap *post processing* dan *tubelet-level bounding box linking* dapat meningkatkan kualitas hasil deteksi, yaitu sekitar 3% lebih baik setelah menggunakan *robust & efficient post processing* dan *tubelet-level bounding box linking*.

Hal ini menunjukkan bahwa *post processing* dan *bounding box linking* merupakan metode yang sesuai untuk *video object detection* sampah di permukaan air karena sifatnya yang dapat memanfaatkan korelasi spasial dan temporal untuk melakukan deteksi pada setiap *frame*. Dengan demikian, penggunaan metode ini berpotensi untuk mendeteksi sampah di perairan dan diaplikasikan pada suatu robot pengumpul sampah otomatis yang dapat mengurangi biaya pengumpulan sampah secara manual yang masih dilakukan hingga saat ini.

## 4.2 Saran

Sebagai negara maritim, Indonesia hendaknya memiliki data video yang menunjukkan kenampakan perairan wilayah Indonesia. Data tersebut dapat digunakan untuk mendeteksi sampah di permukaan perairan sehingga dapat diimplementasikan untuk mendeteksi secara *real time* pada robot otomatis untuk membantu meningkatkan kebersihan perairan di Indonesia. Harapan tim peneliti dengan adanya data tersebut adalah pemerintah terdorong untuk membuat kebijakan terkait pengelolaan sampah di perairan Indonesia, serta program edukasi terkait dampak sampah perairan.

Pengembangan lanjut yang dapat dilakukan adalah melakukan *training* pada dataset Indonesia untuk mendapatkan hasil yang lebih akurat untuk video sampah perairan di Indonesia. Di samping itu, penelitian dapat juga bereksperimen lanjut untuk tidak hanya mengenali sampah plastik, tetapi juga jenis sampah lainnya sehingga dapat lebih bermanfaat untuk meningkatkan kebersihan perairan Indonesia. Selain itu, eksperimen dengan metode deteksi objek video lain yang lebih baru atau memiliki performa yang lebih baik juga dapat diimplementasikan.

Implementasi metode *video object detection* yang disampaikan dalam makalah ini dapat diaplikasikan melalui robot pengumpul sampah otomatis. Implementasi tersebut diharapkan dapat meminimalisasi biaya pemeliharaan dan sumber daya manusia. Dengan demikian, robot ini dapat membantu pemulihan kesehatan masyarakat yang telah disebabkan oleh sampah pada masa lampau dan membuat Indonesia memiliki sistem pengelolaan sampah yang lebih baik pada masa mendatang.

# BIBLIOGRAFI